\PassOptionsToPackage{prologue,dvipsnames}{xcolor}
\documentclass[sigconf]{acmart}

\usepackage{wrapfig}
\usepackage{mathrsfs}
\usepackage{float}
\usepackage{makecell,diagbox}
\usepackage{threeparttable}
\usepackage{tablefootnote}
\usepackage{multirow}

\AtBeginDocument{%
  }
\copyrightyear{2025} 
\acmYear{2025} 
\setcopyright{cc}
\setcctype{by}
\acmConference[CIKM '25]{Proceedings of the 34th ACM International
	Conference on Information and Knowledge Management}{November 10--14,
	2025}{Seoul, Republic of Korea}
\acmBooktitle{Proceedings of the 34th ACM International Conference on
	Information and Knowledge Management (CIKM '25), November 10--14, 2025,
	Seoul, Republic of Korea}\acmDOI{10.1145/3746252.3761045}
\acmISBN{979-8-4007-2040-6/2025/11}

\begin{document}

\title{STA-GANN: A Valid and Generalizable Spatio-Temporal Kriging Approach}

\author{Yujie Li}
\affiliation{%
	\institution{State Key Laboratory of AI Safety, Institute of Computing Technology,\\Chinese Academy of Sciences}
	\country{University of Chinese Academy of Sciences}
	\postcode{100190}
}
\email{liyujie23s@ict.ac.cn}

\author{Zezhi Shao}
\affiliation{%
	\institution{State Key Laboratory of AI Safety, Institute of Computing Technology,}
	\country{Chinese Academy of Sciences}
	\postcode{100190}
}
\email{shaozezhi@ict.ac.cn}

\author{Chengqing Yu}
\affiliation{%
	\institution{State Key Laboratory of AI Safety, Institute of Computing Technology,\\Chinese Academy of Sciences}
	\country{University of Chinese Academy of Sciences}
	\postcode{100190}
}
\email{yuchengqing22b@ict.ac.cn}

\author{Tangwen Qian\\Zhao Zhang\\Yifan Du}
\affiliation{%
\institution{State Key Laboratory of AI Safety, Institute of Computing Technology,}
	\country{Chinese Academy of Sciences}
	\postcode{100190}
}
\email{{qiantangwen, zhangzhao2021}@ict.ac.cn}
\email{duyifan@ict.ac.cn}

\author{Shaoming He}
\affiliation{%
\institution{School of Aerospace Engineering,}
\country{Beijing Institute of Technology}
\postcode{100081}
}
\email{shaoming.he@bit.edu.cn}

\author{Fei Wang}
\authornote{Corresponding author.}
\author{Yongjun Xu}
\affiliation{%
	\institution{State Key Laboratory of AI Safety, Institute of Computing Technology,\\Chinese Academy of Sciences}
	\country{University of Chinese Academy of Sciences}
	\postcode{100190}
}
\email{{wangfei, xyj}@ict.ac.cn}

\renewcommand{\shortauthors}{Yujie Li et al.}

\begin{abstract}
Spatio-temporal tasks often encounter incomplete data arising from missing or inaccessible sensors, making spatio-temporal kriging crucial for inferring the completely missing temporal information. However, current models struggle with ensuring the validity and generalizability of inferred spatio-temporal patterns, especially in capturing dynamic spatial dependencies and temporal shifts, and optimizing the generalizability of unknown sensors. To overcome these limitations, we propose Spatio-Temporal Aware Graph Adversarial Neural Network ~(STA-GANN)~, a novel GNN-based kriging framework that improves spatio-temporal pattern validity and generalization. STA-GANN integrates (i) Decoupled Phase Module that senses and adjusts for timestamp shifts. (ii) Dynamic Data-Driven Metadata Graph Modeling to update spatial relationships using temporal data and metadata; (iii) An adversarial transfer learning strategy to ensure generalizability. Extensive validation across nine datasets from four fields and theoretical evidence both demonstrate the superior performance of STA-GANN.
\end{abstract}



\maketitle
\vspace{-2pt}
\section{Introduction}\label{intro}

Spatio-temporal tasks are crucial in domains such as energy, transportation \cite{zheng2020gman, shao2022pre, shao2022decoupled}, and meteorology \cite{bi2023accurate}, yet sensor unavailability caused by deployment costs, equipment failures often leads to severe data gaps.  To mitigate it, spatio-temporal kriging \cite{wu2021inductive} has become a key technique for inferring completely missing time series through temporal and spatial dependencies of available sensors. 

The primary challenge of spatio-temporal kriging lies in the absence of time series data for certain unknown sensors or nodes. Unlike conventional time series imputation, which utilizes historical and future data to fill in missing values within the same series, spatio-temporal kriging relies on temporal information from other sensors and their spatial relationships to make inferences. As a result, the model must capture spatio-temporal patterns \cite{li2025trajectory} in both a \textbf{valid} and \textbf{generalizable} manner. This involves mining meaningful patterns from available data and ensuring these patterns can be transferred to sensors or nodes without prior observations.

\begin{figure}[t]
	\centering
	\includegraphics[width=0.46\textwidth]{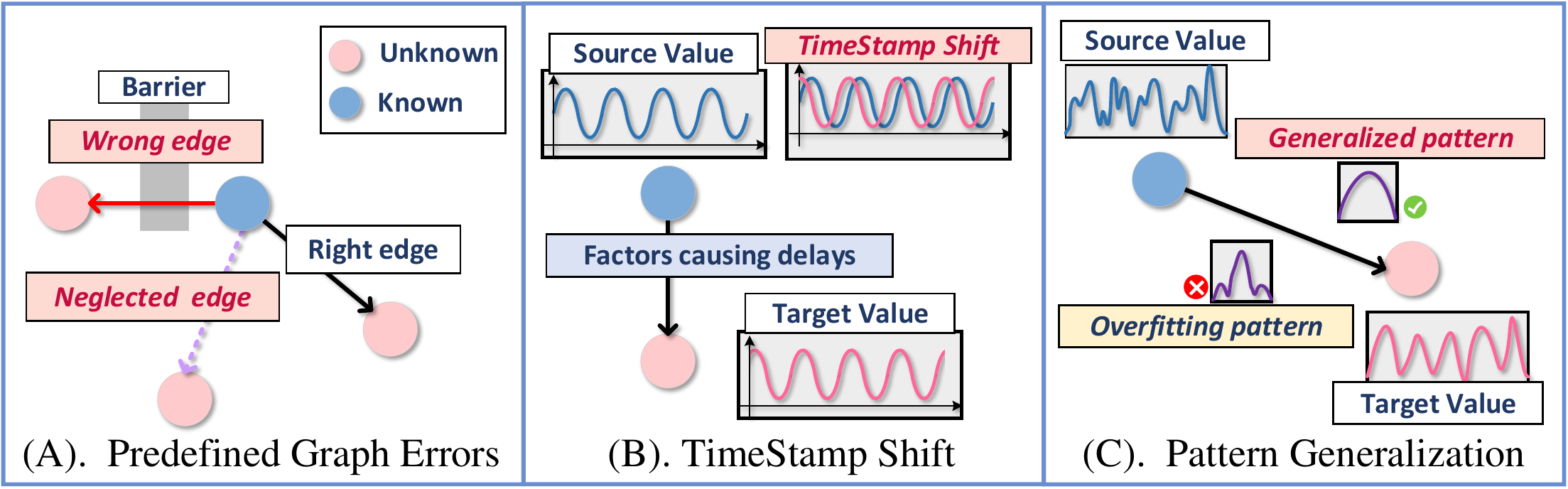}
	\vspace{-7pt}
	\caption{Challenges of Spatio-Temporal Kriging. (A) Predefined graphs may contain unpredictable errors because some edge relations are incorrectly established or ignored. (B) Timestamp shift refers to delays in the transmission of temporal information caused by inevitable factors such as distance and upstream-downstream relationships. 
		(C) The captured overfitted patterns may only be suitable for specific sensors and cannot be generalized.}
	\label{chal}
	\vspace{-10pt}
\end{figure}

Existing methods often struggle to the dual requirements of validity and generalization in spatio-temporal kriging. Traditional approaches, such as Kriging\cite{appleby2020kriging} and matrix decomposition\cite{bahadori2014fast}, typically underperform compared to deep learning models\cite{xu2021artificial}  in real-world scenarios.
IGNNK\cite{wu2021inductive} is the first to adopt graph neural networks\citep{kipf2016semi, gilmer2017neural} for Kriging, establishing the foundation for subsequent research. 
SATCN\cite{wu2021spatial}, IAGCN\cite{wei2024inductive}, and INCREASE\cite{zheng2023increase} all aim to improve performance by extending spatial relationships, but differ in focus: INCREASE leverages scarce metadata including functional areas and points of interest~(POIs), while SATCN and IAGCN focus on optimizing the predefined graph~\citep{yu2018spatio, li2018diffusion} or extracting features from it.
DualSTN\cite{hu2023decoupling} instead targets temporal patterns by disentangling long- and short-term dependencies.
However, these methods still face challenges in capturing valid and generalizable spatio-temporal patterns:
\begin{itemize}
	\item For the validity of spatial patterns, errors in predefined graphs lead to error accumulation during message passing, and current work fails to capture the dynamics of spatial relationships.\citep{wu2019graph, yu2023dsformer, li2023dynamic}.
	\item For the validity of temporal patterns \cite{shih2019temporal, shao2023hutformer}, timestamp shift poses a unique challenge for kriging, which is distinct from other temporal issues like long-term dependence.
	\item For the generalization of spatio-temporal patterns to unknown nodes, Label propagation methods for graph-based semi-supervised learning only guarantee convergence~\cite{iscen2019label}, but overlooks generalization to unknown sensor patterns.
\end{itemize}
\vspace{-2pt}
To address the key challenges of spatio-temporal kriging in terms of validity and generalization, we decompose them into three fundamental challenges and propose corresponding solutions and innovations, as illustrated in Figure~\ref{chal} :

\textbf{1. How to ensure the spatial relationship is appropriate?} \label{challenges}
Spatial relationships are crucial for spatio-temporal kriging. Predefined graphs \citep{yu2018spatio, li2018diffusion}, as shown in Figure~\ref{chal}(A), misrepresent sensor spatial patterns and lack adaptability to environmental changes. This diminishes sensor interactions and causes error accumulation in graph learning, ultimately degrading kriging performance.

OUR APPROACH: We propose \textbf{Dynamic Data-Driven Metadata Graph Modeling~(D3MGM)~} for kriging to capture dynamic spatial dependencies among sensors. This framework leverages temporal data and metadata \citep{li2024dynamic, liu2024largest}, enabling the model to utilize universal information such as coordinates and timestamps for constructing accurate spatial relationships.

\textbf{2. How to learn effective temporal patterns?}
Timestamp shift in Figure~\ref{chal}(B) \citep{niu2023time, li2024dynamic} refers to the signal transmission delay between sensors. This results in the same information being described at different timestamps across sensors, leading to misalignment with target values and degraded metrics. Moreover, this is a unique problem for spatio-temporal kriging, as forecasting relies on historical data and imputation uses contextual information.

OUR APPROACH: We introduce \textbf{Decoupled Phase Module} \textbf{(DPM)} to sense and compensate for timestamp shifts. By adjusting the phase of unknown sensors in the frequency domain \citep{zhou2022fedformer, yi2024frequency} using decoupled series \cite{wu2021autoformer}, we leverage the Fourier Transform's time-shift property \cite{yang2020adaptive} to align timestamps when specific time series information emerges.

\textbf{3. How to ensure that learned spatio-temporal patterns are generalized?} 
During training, unknown sensors are absent, and missing nodes are simulated by randomly masking known sensors. This raises a critical question: how to ensure the model’s generalization to truly unknown sensors and new spatial relationships?

OUR APPROACH:  Despite the importance of inductive GNNs for spatio-temporal kriging, it is still difficult to ensure the generalization of spatio-temporal patterns during label propagation.~\cite{iscen2019label}. To address this, we apply adversarial transfer learning\cite{ganin2016domain} to enhance the generalizability\citep{du2021adarnn, lu2022out}.
We treat known nodes as the source domain and unknown nodes as the target domain. and introduce discriminators which motivates the model to distinguish between known and unknown sensors. If the model successfully distinguishes them, it suggests a parameter state lacking generalizability. Conversely, maintaining strong kriging performance in this scenario indicates robust generalization.

Overall, we propose \textbf{Spatio-Temporal Aware Graph Adversarial Neural Network~(STA-GANN)} to address those problems.
STA-GANN is a spatio-temporal kriging framework that: (1) constructs dynamic graphs to capture dynamic spatial relationships, (2) detects and compensates for timestamp shifts via a decoupled phase module, and (3) employs adversarial transfer learning to enhance generalization.
In the following sections, we detail our approach and demonstrate its effectiveness through theoretical analysis and extensive experiments on multiple real-world datasets. Our contributions are as follows:

\begin{itemize}
	\item We identify and address the dual challenges of validity and generalization in spatio-temporal kriging, proposing the STA-GANN framework.
	\item  We introduce  Dynamic Data-Driven Metadata Graph Modeling and Decoupled Phase Module to perceive valid spatio-temporal patterns.
	\item We pioneer the focus on generalizability in spatio-temporal kriging, leveraging adversarial transfer learning to ensure the performances.
	\item  We conduct extensive experiments on multiple real-world datasets and perform theoretical proofs to demonstrate the effectiveness of our motivations and methods.
\end{itemize}

\section{Related Work}\label{rw}

\subsection{Graph Neural Network}
Graph Neural Networks\citep{hamilton2017inductive, velivckovic2018graph} are deep learning methods for non-Euclidean graphs.
Both GCN\cite{kipf2016semi} and GIN\cite{xu2018powerful} update node representations by aggregating information from neighbors\cite{gilmer2017neural},  with GIN further controlling historical data and performing nonlinear structural aggregation.
Semi-supervised strategies \cite{kipf2016semi}, such as randomly masking labeled nodes for reconstruction, are widely adopted in spatio-temporal kriging to enhance performance. 

\subsection{Spatio-Temporal Forecasting}
Spatio-Temporal forecasting\citep{zhou2021informer, wang2023clustering, yu2023dsformer} is a critical task in multivariate time series analysis, aiming to predict future time series by leveraging historical data and spatial relationships. Traffic prediction \citep{shao2022decoupled, li2018diffusion}, a prominent subfield, focuses more on spatial dependencies and short-term forecasting.
DCRNN\cite{li2018diffusion} utilizes predefined graphs\cite{zheng2019deepstd}, whereas GWNET\cite{wu2019graph} adaptively learns spatial relationships via learnable node embeddings. Recent works further explore dynamic graph learning \cite{li2023dynamic, han2021dynamic, liang2024survey}, which essentially extends GWNET by incorporating dynamic metadata~(e.g., timestamps) to infer spatial dependencies.
However, in spatio-temporal kriging, \textbf{assigning learnable parameters to each node \citep{wu2019graph, shao2022spatial} is infeasible due to scalability constraints.} Thus, effectively modeling dynamic spatial dependencies remains a key challenge in our work.

\subsection{Spatio-Temporal Imputation \& Kriging}
Spatio-temporal imputation\cite{yu2024ginar} aims to fill missing values within known nodes, with the primary challenge being the randomness of missing values. However, due to scalability limitations, only a few works such as GRIN\cite{cini2021filling} are adaptable to kriging after modifications. 
IGNNK\cite{wu2021inductive} is the first deep kriging model, through GCN and residual network to achieve a good performance of spatio-temporal kriging. SATCN \cite{wu2021spatial} enhances spatial information by exploring predefined graph variants, and DualSTN\cite{hu2023decoupling} concerns the analysis of long-term and short-term temporal patterns. Additionally, INCREASE\cite{zheng2023increase} leverages metadata such as POIs and functional areas to construct diverse predefined graphs and prioritizes nearest-neighbor sensors during aggregation. IAGCN \cite{wei2024inductive} expands spatial relations by re-learning from known adjacency matrices.

\subsection{Transfer Learning}
Transfer learning\citep{ghifary2014domain, long2017deep, zhuang2020comprehensive} aims to transfer knowledge from a source domain to a target domain, typically across distinct data distributions. In spatio-temporal kriging, this aligns with transferring patterns from known sensors to unknown sensors.
Adversarial transfer learning\cite{saito2018maximum, sun2016deep2}, exemplified by Domain-Adversarial Neural Networks~(DANN)\cite{ganin2016domain},employs adversarial training to learn domain-invariant feature representations. This framework typically involves a feature extractor and a domain discriminator: the extractor aims to confuse the discriminator \cite{goodfellow2014generative}, minimizing domain discrepancy to achieve generalization by preventing the discriminator from distinguishing features of the source and target domains.

\section{Proposed Method}\label{pm}
In this section, we provide an overview of STA-GANN in the order of data flow, and focus on the solutions to three challenges of spatio-temporal kriging presented in Section~\ref{challenges} , including decoupled phase module~(DPM)~, dynamic data-driven metadata graph modeling~(D3MGM)~, and adversarial transfer learning strategies.

\textbf{Problem Definition} Let \( G_K \!=\! \{V_K,\! E_K,\! A_K,\! M_K,\! X_K\} \) denote the graph of \( N_K \) known sensors, where \( A_K \) encodes spatial relationships, \( M_K \) contains metadata such as coordinates and timestamps, and $X_{K}=\{ X_{i}\}^{N_{K}}_{i=1}$ with  $X_{i}=\{x_{i,t}\}^{T}_{t=1}$ is \( T \)-length time series.

Spatio-temporal kriging model $\mathcal{F}$ first learns to reconstruct $\tilde{X_K} = \mathcal{F}(X_K, G_K)$ during training. For unseen sensors $ G_U = \{V_U, E_U$, $A_U, M_U\}$~(Completely missing $X_U$), it infers $X_U \!=\! \mathcal F (X_K, G_K + G_U)$ by leveraging learned spatio-temporal dependencies during validation and testing.

\subsection{Overview}
\begin{figure*}
	\centering
	\includegraphics[width=0.8\textwidth]{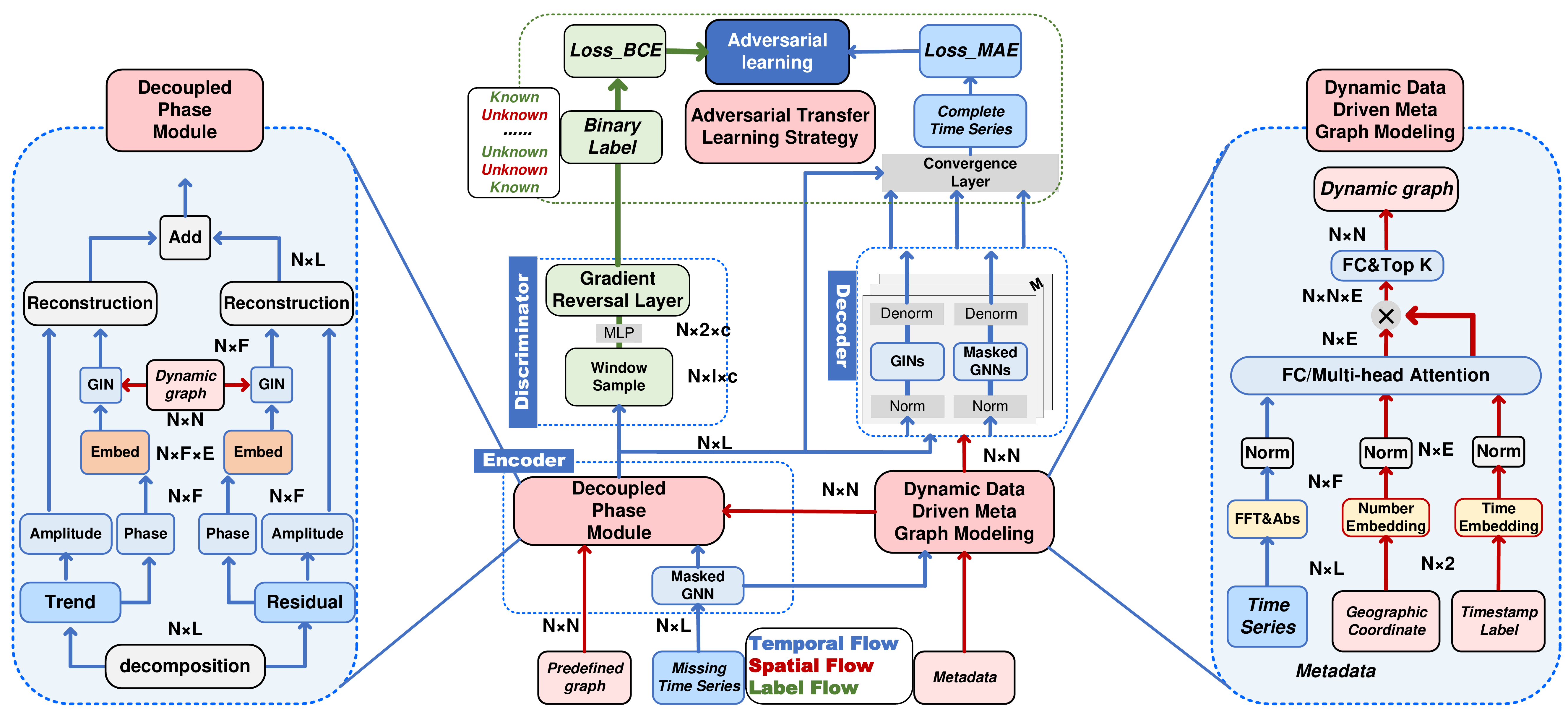}
	\vspace{-5pt}
	\caption{Architecture of STA-GANN}
	\vspace{-5pt}
	\label{stagann}
\end{figure*}

As depicted in Figure~\ref{stagann}, STA-GANN receives:
\begin{itemize}
	\item Time series $X\in R^{N \times L}$, where unknown sensors initialized as $X_{U}=0$,
	\item Adjacency matrix $A\in R^{N \times N}$ encoding spatial relationships,
	\item Metadata $M$,
\end{itemize}
We distinguish the data flows with different colors and emphasize that the number of sensors $N$ is arbitrary.

\textbf{In the encoder}, time series first passes through Masked GNN to obtain temporal information, then enters the Decoupled Phase MModule~(DPM) to learn timestamp shifts between sensors. We adopt the graph isomorphism network~(GIN) as the backbone, The masked GNN resembles GCN but removes GIN's historical message filtering parameters since unknown sensors lack historical data.
The hidden layer dimension of MLPs  is denoted as $H$, the superscript \( k\) indicates the \( k \)-th MLP layer, and \( s \) is a learnable parameter in GIN that controls the strength of history information. The operations of Masked GNN and GIN are defined as follows:
\begin{flalign}\label{gcn1} 	 	
	X_{i}^{(k)} \!= MLP^{(k)}\!(\sum_{j\in V} e_{i,j}\!\cdot\! X_{j}^{(k-1)}\!) \!=\! MLP^{(k)}\!(AX^{(k-1)})
\end{flalign}
\vspace{-8pt}
\begin{flalign}\label{gcn2} 
	X_{i}^{(k)} \!&= MLP^{(k)}\!((1\!+\!\epsilon^{(k-1)}\!)\!\cdot\! X_{i}^{(k-1)} \!+\! \sum_{j\in V} e_{i,j}\!\cdot\! X_{j}^{(k-1)}\!) \notag\\
	\!&= MLP^{(k)}\!((1\!+\!\epsilon^{(k-1)})\!\cdot\! X_{i}^{(k-1)}+ AX^{(k-1)})
\end{flalign}

In Masked GNN, predefined graphs are the only source of spatial dependencies. In contrast, the Decoupled Phase Module~(DPM) utilizes adjacency matrices generated by Dynamic Data-Driven Metadata Graph Modeling~(D3MGM), which derives spatial relationships from temporal information.

For DPM, we highlight its role in sensing and compensating for timestamp shifts during message passing by leveraging frequency-domain information and sensor spatial relationships. 

\textbf{In the decoder}, the encoder’s output serves as the initial layer input, with subsequent layers connected sequentially. In Figure~\ref{stagann}, the decoder employs a dual-data flow structure, differing in whether the graph network prioritizes sensor self-information. We utilize Masked GNN and GIN, respectively, allowing the convergence layer to adaptively filter dataset-specific information and mitigate error accumulation, a common issue in graph networks. Furthermore, we apply Revin\cite{kim2021reversible} to normalize\cite{liu2022non} inputs and inverse-normalize outputs at each layer, reducing the impact of distributional drift.

Furthermore, all GNN outputs preserve the original input dimensions, despite their hidden layer dimension $H$ of the MLPs. Additionally, each $N \times L$-dimensional vector restores the original information of known sensors during layer-wise propagation and is residually connected to the convergence layer. The convergence layer consists of multiple convolutional layers and performs channel-wise weighting to produce the final output.

\textbf{In the discriminator}, it takes the encoder’s output as input. The discriminator is a simple MLP designed to classify whether the tensors come from known or unknown sensors. Given the limited number of time series,  we construct a sliding window of length $l$, smaller than the time series length, to slice the time series into shorter patches. These patches are then fed into the discriminator for binary classification.
\vspace{-10pt}
\subsection{Dynamic Data-Driven Metadata Graph Modeling}
To address the unreliability of predefined graphs, we propose Dynamic Data-Driven Metadata Graph Modeling~(D3MGM) to capture dynamic spatial relationships in spatio-temporal kriging. Unlike approaches that assign learnable node parameters in forecasting, D3MGM derives spatial relations directly from temporal data, enriched by metadata including timestamps $Label_{time}$ and coordinates $Label_{coor}$ due to the node variability constraints of kriging.

For timestamps, we follow standard practice in short-term forecasting and construct an index-based lookup table $Emb_{time}$ based on time-of-day and day-of-week, for example, to achieve effective representation of time metadata.

For coordinates, prior methods often depend on dataset-specific preprocessing~(e.g., POIs or grid indexing \cite{zheng2021stpc, liang2024mgksite}). Instead, we directly embed numerical coordinates by initializing a shared number embedding $Emb_{number}\in R^{N\times E}$ from 0 to 9. Each coordinate is tokenized into digits, and their embeddings are concatenated to form $Emb_{coor}\in R^{N\times E}$, which aligns with the tokenization strategy of large language models like LLaMa for numerical data processing.

Temporal signals are further transformed into the frequency domain $FFT(X)\in R^{N\times F}$ following DFDGCN \cite{li2024dynamic}. Sensor embeddings are then generated via either an attention-based or linear approach, depending on the availability of coordinate information:
\begin{flalign}\label{DE}  	 	
	DE \!&=\! MHA((Emb_{time}||Emb_{coor}), FFT(X), FFT(X)) \\
	DE \!&=\! Linear(Concat(Emb_{time}, FFT(X)))
\end{flalign}

To construct the graph structure, we employ the embedding $DE \in R^{N\times E}$ to learn the graph structure. Instead of assigning learnable parameters to $DE$ and its transpose to capture directionality, we introduce a learnable parameter $W_{e}\in R^{E\time 1}$ to model edge information along dimension $E$. Finally, the adjacency matrix $A_{gm}$ is generated by selecting the $TopK$ neighbors \cite{wu2020connecting} for each sensor:
\begin{flalign}\label{topk} 	 	
	A_{gm} = TopK(W_{e}(DE \otimes DE^{T}))
\end{flalign}

\subsection{Decoupled Phase Module}
Decoupled Phase Module~(DPM) is employed to sense and complement the timestamp shifts between sensors. The general idea is to decouple the trend and residual components of the time series, allowing the model to learn phase shifts under different components.

\textbf{Time series decoupling}. Time series generally comprise periodic and trend components \cite{huang1998empirical, zhou2022fedformer, liang2023learn}.  Decoupling these components mitigates feature entanglement and facilitates effective temporal modeling.  We adopt a standard decomposition method, using average pooling convolutions to extract the trend $X^{t}$ and residual~(periodic) term $X^{r}$:
\begin{flalign}\label{avg} 	 	
	X^{t} = Average(Pooling(X))
\end{flalign} 
\vspace{-15pt}
\begin{flalign}\label{xxt} 	 	
	X^{r} = X - X^{t}
\end{flalign}
Furthermore, we employ the Fourier Transform to convert time-domain information into the frequency domain, emphasizing amplitude and phase information that are essential for sequence reconstruction. Since timestamp shifts strongly correlate with phase variations, we primarily focus on modeling phase shifts.

The rationale for learning shifts in the frequency rather than time domain lies in its efficiency and reduced risk of overfitting. In this domain, we only model how one phase shifts relative to another based on spatial relationships and their corresponding embeddings. This approach prevents deep learning models from struggling to isolate timestamp shifts and learning irrelevant temporal semantics, which could otherwise potentially lead to overfitting and negatively impact the model's generalization capability and performance.

\begin{figure}[h]
	\centering
	\includegraphics[width=0.42\textwidth]{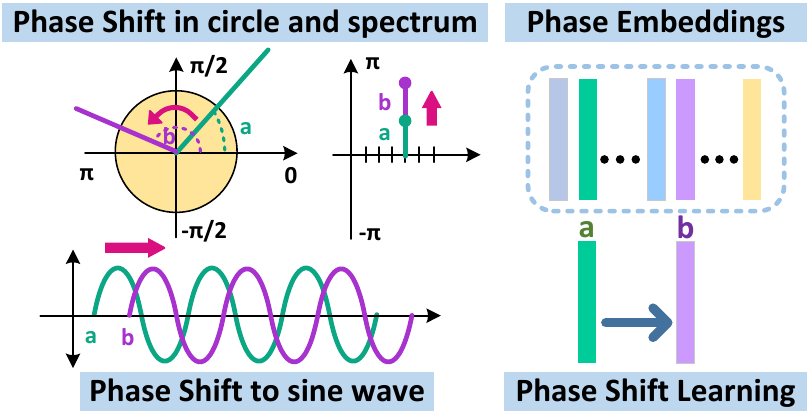}
	\vspace{-10pt}
	\caption{Phase Shift Learning}
	\label{phase}
	\vspace{-10pt}
\end{figure}
\textbf{Phase Embedding:} As shown in Figure~\ref{phase} , since the phase range is circular, shifting the phase of the frequency components within this range causes the corresponding time-domain sinusoid to be shifted in the time-stamped dimension. We discretize the range $[-\pi,\pi]$ into $M$ segments, each with a separate embedding, so any phase $\phi_i$ is represented by an embedding vector $\vec{\phi_{i,m}}$.

D3MGM constructs the graph $A_{\phi}$ during this learning process, enabling specialized GIN to learn the phase shifts effectively. Through GIN, interactions between phase embeddings of each frequency domain component within sequences are facilitated.  Ultimately, a fully connected layer transforms high-dimensional phase embeddings into a single one-dimensional phase at the final $K$-th layer output. Notably, the phase embedding $\vec{\phi_{i,m}}$ remains unlearnable~(see Equation~\ref{mlpphase}), while other parameters are trainable Since it only represents natural information from $-\pi$ to $\pi$, random initialization is sufficient to maintain the distinguishability between phase vectors:
\begin{flalign} 	 	
	\vec{\phi_{m}} \!=\! Embedding(\phi_{i})
\end{flalign}
\vspace{-10pt}
\begin{flalign}\label{mlpphase} 	 	
	\vec{\phi_{i,m}}^{k} \!=\! MLP((1\!+\epsilon^{(k-1)})\cdot \vec{\phi_{i,m}}^{(k-1)}\!+\! A_{phase}\vec{\Phi}^{(k-1)})
\end{flalign}
\vspace{-10pt}
\begin{flalign} 	 	
	\phi_{i}^{K} = tanh(FC(\vec{\phi_{i,m}}^{K})) \!\cdot\! \pi
\end{flalign}
Upon securing the new phases, we utilize them to reconstruct the time series by performing an inverse Fourier Transform with amplitudes. The trend and residual terms are each subjected to this process independently. Once reconstructed, the time series will be aggregated as the output of DPM and subsequently fed into both the decoder and discriminator components of the system.

\textbf{Theoretical explanation: }The time-shift property of Fourier Transform~\cite{demirel2025shifting} establishes that temporal translation corresponds to phase rotation in the frequency domain. Let \( f(t) \) be a signal with Fourier Transform \( F(\omega) \). For any time shift \( t_0 \), the shifted signal \( f(t \pm t_0) \) satisfies:
\begin{flalign}
	\mathcal{F}[f(t \pm t_0)] &= e^{\pm j\omega t_0}F(\omega)
\end{flalign}

Generalizing for left/right shifts (\( \pm t_0 \)) and substituting \( \phi = \omega t_0 \):
\begin{flalign}
	\mathcal{F}[f(t - t_0)] &= e^{-j\phi}F(\omega) \\
	\mathcal{F}[f(t + t_0)] &= e^{j\phi}F(\omega)
\end{flalign}

Applying Euler's formula \( e^{\pm j\phi} = \cos\phi \pm j\sin\phi \) and Neglecting imaginary components:
\begin{flalign}
	\mathcal{F}[f(t \pm t_0)] = [\cos\phi \pm j\sin\phi]F(\omega) &\approx \cos\phi \cdot F(\omega)
\end{flalign}
The DPM utilizes a simple transformation of time shift into the frequency domain to learn the phase changes \( \phi \in (-\pi, \pi] \) in graphs.

\begin{table*}[h]
	\caption{Results of models on nine real datasets. The best results are in \textbf{bold}, and the second best results
		are \underline{underlined}.}
	\vspace{-5pt}
	\label{result}
	\centering
	\renewcommand\arraystretch{0.6}
	\resizebox{0.91\linewidth}{!}{
	\begin{tabular}{cp{25pt}|p{45pt}|p{35pt}|p{35pt}|p{35pt}|p{35pt}|p{30pt}|p{30pt}|p{30pt}|p{30pt}|p{30pt}}
		\toprule[1.5pt]
		Datasets & Metrics& STA-GANN&GRIN&INCREASE&DualSTN&IAGCN&IGNNK&SATCN&GCN&GIN&Okriging\\
		\midrule
		\multicolumn{1}{c}{\multirow{3}{*}{METR-LA}}  & \textit{MAE}&\textbf{4.887}&5.209&\underline{5.121}&5.9165&6.2377&5.770&5.428&7.68&6.253&8.393\\
		\multicolumn{1}{c}{}& \textit{RMSE}&\textbf{8.012}&\underline{8.543}&8.640&10.228&10.095&9.436&8.792&10.355&9.529&10.959\\
		&\textit{$R^{2}$}&\textbf{0.861}&\underline{0.842}&0.795&0.774&0.780&0.808&0.636&0.768&0.804&0.741\\
		\midrule
		\multicolumn{1}{c}{\multirow{3}{*}{PEMS-BAY}} &\textit{MAE}&\textbf{3.673}&\underline{3.708}&3.739&3.921&4.399&4.128&4.337&3.993&4.307&4.116\\
		\multicolumn{1}{c}{}&\textit{RMSE}&\textbf{6.495}&\underline{6.716}&6.793&7.152&7.804&7.424&7.968&7.061&8.181&7.247\\
		&\textit{$R^{2}$}&\textbf{0.593}&\underline{0.566}&0.558&0.507&0.413&0.384&0.389&0.520&0.357&0.494\\
		\midrule
		\multicolumn{1}{c}{\multirow{3}{*}{PEMS03}} &\textit{MAE}&\textbf{64.064}&\underline{70.044}&75.155&75.151&83.161&77.210&73.703&87.816&82.676&77.760\\
		\multicolumn{1}{c}{}&\textit{RMSE}&\textbf{99.443}&\underline{105.447}&116.561&110.662&123.881&115.961&105.722&118.303&117.370&108.324\\
		&\textit{$R^{2}$}&\textbf{0.576}&\underline{0.548}&0.379&0.440&0.298&0.385&0.489&0.360&0.370&0.464\\
		\midrule
		\multicolumn{1}{c}{\multirow{3}{*}{PEMS04}} &\textit{MAE}&\textbf{59.011}&\underline{61.256}&65.264&61.695&64.848&65.295&76.038&84.149&69.701&69.713\\
		\multicolumn{1}{c}{}&\textit{RMSE}&\textbf{81.992}&\underline{84.566}&91.229&85.329&86.703&90.316&98.956&111.805&89.864&95.659\\
		&\textit{$R^{2}$}&\textbf{0.674}&\underline{0.655}&0.587&0.639&0.627&0.596&0.515&0.381&0.599&0.546\\
		\midrule
		\multicolumn{1}{c}{\multirow{3}{*}{PEMS07}} &\textit{MAE}&\textbf{69.302}&70.779&74.002&80.246&74.639&\underline{69.604}&88.491&100.749&102.185&83.767\\
		\multicolumn{1}{c}{}&\textit{RMSE}&\textbf{100.190}&104.515&107.955&116.376&107.282&\underline{102.338}&127.582&132.147&134.694&118.140\\
		&\textit{$R^{2}$}&\textbf{0.71}&0.685&0.664&0.610&0.668&\underline{0.698}&0.531&0.490&0.478&0.598\\
		\midrule
		\multicolumn{1}{c}{\multirow{3}{*}{PEMS08}}&\textit{MAE}&\textbf{86.329}&92.291&94.833&93.867&92.145&97.072&93.241&94.014&95.639&\underline{91.177}\\
		\multicolumn{1}{c}{}&\textit{RMSE}&\textbf{117.203}&123.778&124.515&123.573&123.564&135.645&122.677&122.789&125.459&\underline{122.092}\\
		&\textit{$R^{2}$}&\textbf{0.417}&0.351&0.343&0.352&0.353&0.220&\underline{0.362}&0.361&0.333&0.358\\
		\midrule
		\multicolumn{1}{c}{\multirow{3}{*}{NREL}} &\textit{MAE}&\textbf{3.965}&\underline{4.212}&4.638&4.415&4.614&4.509&4.417&5.927&4.989&6.508\\
		\multicolumn{1}{c}{}&\textit{RMSE}&\textbf{6.851}&7.153&7.444&\underline{7.144}&7.219&7.267&7.106&8.859&7.694&9.471\\
		&\textit{$R^{2}$}&\textbf{0.430}&\underline{0.394}&0.327&0.380&0.367&0.359&0.387&0.047&0.281&-0.089\\
		\midrule
		\multicolumn{1}{c}{\multirow{3}{*}{USHCN}} &\textit{MAE}&\textbf{2.190}&2.832&\underline{2.248}&3.912&3.769&2.793&2.767&3.136&3.746&3.158\\
		\multicolumn{1}{c}{}&\textit{RMSE}&\textbf{3.551}&4.384&\underline{3.771}&5.909&5.723&4.312&4.260&4.708&5.663&4.708\\
		&\textit{$R^{2}$}&\textbf{0.711}&0.559&\underline{0.674}&0.200&0.250&0.572&0.564&0.492&0.265&0.492\\
		\midrule
		\multicolumn{1}{c}{\multirow{3}{*}{AQI}} &\textit{MAE}&\textbf{15.361}&16.727&\underline{16.626}&20.213&27.812&18.162&16.845&23.660&26.478&23.201\\
		\multicolumn{1}{c}{}&\textit{RMSE}&\textbf{27.935}&29.944&\underline{29.695}&34.249&42.544&32.381&30.720&37.498&40.919&36.822\\
		&\textit{$R^{2}$}&\textbf{0.588}&0.528&\underline{0.539}&0.382&0.046&0.448&0.503&0.260&0.119&0.286\\
		\bottomrule[1.5pt]
		
	\end{tabular}
	}
\end{table*}

\subsection{Adversarial Transfer Learning Strategy}
As previously detailed, we apply a sliding window to segment the temporal data $X_{encoder}\in R^{N\times L}$ from the encoder into shorter patches $X_{patch}\in R^{N\times l\times c}$, where $L$ refers to the window length of each input and $l$ is a smaller patch length. Subsequently, an MLP-based discriminator is utilized to generate a binary classification label $Label_{D} \in R^{N\times 2\times c}$, as formulated below:
\begin{flalign} 	 	
	Label_{D} \!=\! MLP(Window(X_{encoder}))
\end{flalign}
The purpose of $Label_{D}$ is to identify whether the temporal data originates from a known or unknown node. Following the application of the Gradient Reversal Layer~(GRL), the discriminator's loss function decreases while the encoder optimizes in an opposing direction. In this adversarial training framework, the discriminator seeks to distinguish between known and unknown nodes, whereas the encoder aims to obscure these distinctions. 

In our model, both the discriminator’s binary cross-entropy loss $Loss_{D}$ and Kriging's mean absolute error~(MAE) loss $Loss_{main}$ are jointly optimized during the initial epochs, where $Loss_{main}$ is computed only on missing nodes to ensure the model focuses on reconstructing unavailable sensor data.  
Due to the instability of adversarial training,  we inject a small fraction of noise into the real labels $Label_{real}$ for discriminative loss while ensuring that the main loss remains the dominant gradient signal throughout training.
The formulation of the loss functions for the initial training stage is as follows:
\begin{flalign} 	 	
	Loss =& Loss_{D} \!+\! Loss_{main} \\ \notag
	=& BCE(Label_{D}, Label_{real}) \!+\! Loss_{main}
\end{flalign}
After this stage, the discriminator is frozen.

\textbf{Theoretical explanation: } The theory of our approach~\cite{ben2010theory} is from the classical Generalization Error Bound in the target domain based on $\mathcal{H}$-divergence for transfer learning. For hypothesis class $\mathcal{H}$ with VC-dim $d$, with probability $1-\delta$, $\forall h \in \mathcal{H}$:
\begin{flalign} \label{Hdiv}
	R_{\mathcal{D}_t}(h) \!\le\! R_s(h)\! +\! \hat{d}_{\mathcal{H}}(\mathcal{D}_s,\mathcal{D}_t) \!+\! \lambda^{*} 
	\!+\! \sqrt{\frac{4}{n}(d \log \frac{2en}{d} + \log \frac{4}{\delta})} 
\end{flalign}

The bound contains four components: (1) empirical source error $R_s(h)$, (2) domain discrepancy $\hat{d}_{\mathcal{H}}$, (3) irreducible term $\lambda^*$, and (4) complexity term. As $\lambda^*$ is intractable with unlabeled target domains, we focus on minimizing $R_s(h)$~($Loss_{main}$) and $\hat{d}_{\mathcal{H}}$~($Loss_{D}$). 

Given the known sensor domain $\mathcal{D}_s$ and unknown sensor domain $\mathcal{D}_t$, $X$ is time series defined over them and a hypothesis class $\mathcal{H}$. Then the $\mathcal{H}$-divergence between the two domains $\mathcal{D}_s,\mathcal{D}_t$ is defined as: 
\begin{flalign}
	\hat{d}_{\mathcal{H}}(\mathcal{D}_s,\! \mathcal{D}_t) \!=\! 2 \!\sup_{\eta \in \mathcal{H}} \left|\underset{\mathbf{x} \in \!\mathcal{D}_s}{P}[\eta(\mathbf{x}) \!=\! 1] \!-\!\!\! \underset{\mathbf{x} \in \!\mathcal{D}_t}{P}[\eta(\mathbf{x}) \!=\! 1] \right|
\end{flalign}
\begin{flalign}
	\hat{d}_\mathcal{H} (\mathcal{D}_s, \mathcal{D}_t) = 2 &\left\{ 1 - \min_{\eta \in \mathcal{H}} \left[ \frac{1}{n_1} \sum_{i=1}^{n_1} I[\eta(\mathbf{x}_i) = 0] \right. \right. \notag \\ 
	& \quad \left. \left. + \frac{1}{n_2} \sum_{i=1}^{n_2} I[\eta(\mathbf{x}_i) = 1] \right] \right\}
\end{flalign}

Optimizing $\hat{d}_{\mathcal{H}}$ turns out to minimize the maximization of the discriminative loss, which we achieve through adversarial learning:
\begin{itemize}
	\item \textit{Max Step:} Randomly mask known sensors and mislabel them as "unknown" to confuse domains.
	\item \textit{Min Step:} Optimize BCE loss (Eq.12) to align distributions.
\end{itemize}
This minmax strategy bridges $\mathcal{H}$-divergence theory with kriging generalization, analogous to DANN's adversarial mechanism~\cite{ganin2016domain}.

\section{Experiments}\label{exper}

\subsection{Datasets}
\begin{table}[t]
	\centering
	\renewcommand\arraystretch{0.6}
	\setlength{\tabcolsep}{2pt}
	\caption{Summary of Datasets}
	\vspace{-5pt}
	\label{datasets}
	\resizebox{0.95\linewidth}{!}{
		\begin{tabular}{lllll}
			\toprule[1.5pt]
			\textbf{Domain} & \textbf{Datasets} & \textbf{Sensors} & \textbf{TimeSteps} & \textbf{Timestamp} \\
			\midrule
			Traffic Speed & METR-LA\cite{li2018diffusion} & 207 & 34,272 & Mar 2012 - Jun 2012 \\
			& PEMS-BAY\cite{li2018diffusion}  & 325 & 52,116 & Jan 2017 - Jun 2017 \\
			\midrule
			Traffic Flow & PEMS03 & 358 & 26,208 & Sep 2018 - Nov 2018 \\
			& PEMS04 \cite{guo2021learning}  & 307 & 16,992 & Jan 2018 - Feb 2018 \\
			& PEMS07 & 883 & 28,224 & May 2017 - Aug 2017 \\
			& PEMS08 & 170 & 17,856 & Jul 2016 - Aug 2016 \\
			\midrule
			Energy & NREL\cite{wu2021inductive}  & 137 & 105,120 & 2006 \\
			Weather & USHCN\cite{wu2021inductive}  & 1,218 & 1,440 & 1899 - 2019 \\
			Environment & AQI\cite{cini2021filling}  & 437 & 59,710 & Jan 2015 to Dec 2022 \\
			\bottomrule[1.5pt]
		\end{tabular}
	}
	\vspace{-10pt}
\end{table}

\textbf{Datasets}: We evaluate STA-GANN on nine real-world datasets across four domains in Table ~\ref{datasets}. Sensors are split into training~(Known), validation~(Unknown), and testing~(Unknown) sets in a 7:1:2 ratio, while time steps are partitioned 7:3 for trained and untrained intervals. Temporal data and metadata of unknown sensors are unavailable during training. 

\textbf{Baselines and Metrics}: STA-GANN is compared against: (1) pure GNNs such as GCN \cite{kipf2016semi} and GIN \cite{xu2018powerful}; (2) spatio-temporal kriging models like IGNNK \cite{wu2021inductive}, SATCN \cite{wu2021spatial}, INCREASE \cite{zheng2023increase}, DualSTN\cite{hu2023decoupling} and IAGCN\cite{wei2024inductive} ; (3) GRIN\cite{cini2021filling}, a SOTA imputation model adapted for kriging; (4) Parameter-free Okriging aggregating 1st-order neighbor messages.
Performance is measured by Mean Absolute Error~(MAE), Root Mean Square Error~(RMSE), and the coefficient of determination~$R^{2}$.

\textbf{Hyperparameters}: Table~\ref{para} summarizes the main hyperparameters, with others following baseline defaults. Series-level embedding such as STA-GANN are used with temporal dimension set to 100, while timestep-based embeddings use a dimension of 16; SATCN exclusively uses a predefined random walk graph, while STA-GANN and others typically rely on the native dataset graphs.

Our Kriging process is based on the spatio-temporal benchmark BasicTS~\cite{shao2024exploring}, our code is publicly available to facilitate reproducibility and further research: \url{https://github.com/blisky-li/STAGANN}.

\begin{table}[t]
	\centering
	\caption{Summary of Parameters}
	\vspace{-5pt}
	\label{para}
	\renewcommand\arraystretch{0.6}
	\footnotesize
	\resizebox{0.95\linewidth}{!}{
	\begin{tabular}{ll}
		\toprule[1.5pt]
		\textbf{Parameter Name} & \textbf{Description} \\
		\midrule
		Length of Time Series $L$ & 24 \\
		Number of Training Rounds & 50  \\
		Batch Size & 64  \\
		Optimizer & Adam \\
		Loss & MAE (+ our BCE) \\
		Metadata Embedding Dimension $F$ & 20  \\
		Adversarial Strategy Rounds & 5 \\
		Label Embedding Dimension $E$ & 12  \\
		Dropout Rate & 0.3 \\
		\midrule
		Temporal Embedding Dimension $H$ & \makecell[l]{100 (IGNNK, GNNs, DualSTN, \\ IAGCN, STA-GANN) \\ 16 (others)} \\
		\midrule
		Learning Rate & \makecell[l]{0.003 (Most) \\ 0.01 (GRIN, INCREASE)} \\
		\midrule
		Predefined graph & \makecell[l]{Original (Most) \\ Random-Walk (SATCN)}\\
		\midrule
		TopK(ours) & \makecell[l]{50 (PEMS0X)\\ 5 (others)}\\
		\bottomrule[1.5pt]
	\end{tabular}
	}
		\vspace{-10pt}
\end{table}

\begin{table*}[t]
	\caption{Ablation study of STA-GANN, italics indicate variants that have lost the corresponding module. The best results are in bold, and the second best results.
		are underlined.}
	\vspace{-4pt}
	\label{result2}
	\footnotesize
	\centering
	\renewcommand\arraystretch{0.5}
	\setlength{\tabcolsep}{2pt}
	\resizebox{0.95\linewidth}{!}{
	\begin{tabular}{cc|c|c|c|c|c|c|c|c|c|c|c}
		\toprule[1.5pt]
		Datasets & Metrics & \scriptsize \textbf{STA-GANN} & \scriptsize \textit{STA-GANN-S} &\scriptsize \textit{\makecell[c]{-S-location}} &\scriptsize \textit{\makecell[c]{-S-timestamp}}& \scriptsize \textit{STA-GANN-T}&\scriptsize \textit{\makecell[c]{-T-phasegraph}} &\scriptsize \textit{\makecell[c]{-T-decouple}}& \scriptsize \textit{STA-GANN-A} & \scriptsize \textit{-A10}& \scriptsize \textit{-A50} & \scriptsize \textit{\makecell[c]{STA-GANN\\-Revin}} \\
		\midrule
		\multirow{2}{*}{METR-LA} & MAE & \textbf{4.887} & 5.122 & 4.948&5.095 &5.068 & 5.069&4.973& 5.235&4.962&4.928 & \underline{4.914}\\
		& RMSE & \underline{8.012} & 8.306 &\textbf{8.003}& 8.048&8.195 &8.170&8.112&8.353&8.138&8.089 & 8.032\\
		\midrule
		\multirow{2}{*}{PEMS08} & MAE & \textbf{86.329} & 91.122 &-&89.706& 91.692 &90.362&87.850& 89.326&\underline{87.623}&87.723 & 87.842\\
		& RMSE & \textbf{117.203} & 120.899 &-& 119.568&121.116 &120.025&118.239& 120.261 &\underline{117.624}&117.895 & 117.982\\
		\midrule
		\multirow{2}{*}{NREL} & MAE & \underline{3.965} & 4.191 &3.976& 4.125& 4.065 &4.145 &4.009& 4.062&\textbf{3.927}&4.020 &3.972\\
		& RMSE & 6.851 & 6.924 &\underline{6.848}& 6.916&6.889 &6.968&6.865& 6.902 &\textbf{6.780}&6.863&6.823\\
		\midrule
		\multirow{2}{*}{USHCN} & MAE & \underline{2.190} & 2.267 &2.217&2.241& 2.215 &2.215 &2.204& 2.230 &2.205&\textbf{2.183}&2.197\\
		& RMSE & \textbf{3.551} & 3.810 &3.750&3.706& 3.638 &3.784&3.613& 3.696&3.633&\underline{3.587} & 3.623 \\
		\bottomrule[1.5pt]
	\end{tabular}}
\end{table*}

\subsection{Main Results}
As shown in Table ~\ref{result}, STA-GANN achieves state-of-the-art performance across nine real-world datasets, with consistent improvements of over 5\% compared to existing baselines. 
Particularly, it not only surpasses spatio-temporal kriging methods like INCREASE, DualSTN and IAGCN but also achieves nearly a 10\% improvement on datasets such as PEMS0X, NREL, and AQI. This enhanced performance is attributed to the model's ability to validly capture spatio-temporal patterns between sensors and generalize these patterns to unknown sensors. The unmodified GIN and GCN exhibit performance comparable to OKriging in the Kriging problem. Regarding graph learning methods, both SATCN and IAGCN operate on predefined graphs, which amplifies error accumulation. While DualSTN, INCREASE, and GRIN demonstrate improvements in temporal learning, their performance remains constrained by unreliable pre-defined graph . Furthermore, lacking strategies to enhance generalization capability, these methods may achieve good performance on certain datasets but suffer from significant performance degradation in others due to overfitting issues.

\subsection{Ablation Study}
To assess the contributions of STA-GANN’s components, we perform ablation studies by progressively removing key modules and strategies. Notation \textit{"-S/-T/-A"} indicates the removal of D3MGM, DPM, and adversarial transfer learning strategies. Furthermore, \textit{-S -location/-timestamp} remove of location or timestamp information in Equation~\ref{DE}; \textit{-T -phasegraph/-decouple} replaces dynamic graphs in Equation~\ref{mlpphase} with predefined graphs, and removes time-series decoupling from Equations~\ref{avg} and \ref{xxt}; \textit{-A10/A50} freezes the discriminator after 10 rounds or applies no freezing.

We evaluate on four representative datasets~(METR-LA, PEMS08, NREL, and USHCN) covering diverse scenarios. Additionally, we remove Revin in the decoder, which shows negligible impact.

As shown in Table~\ref{result2}, STA-GANN consistently outperforms its ablation variants. Notably, Removing D3MGM in \textit{STA-GANN-S} causes a performance drop exceeding 4\%, highlighting its critical role in enhancing model effectiveness. Both location and timestamp information are essential for accurate graph modeling, with timestamps being particularly influential. These results confirm that predefined graphs often contain inherent biases, while dynamic spatial dependencies captured by D3MGM effectively mitigate such limitations.

From the results in Table~\ref{result2}, both DPM and adversarial transfer learning demonstrate benefits, though they show sensitivity to specific data patterns.
DPM requires stable periodicity and timestamp shifts for improvements, and the \textit{"-phasegraph"} also confirms the importance of spatial relationships, and time-series decoupling largely has the effect of maintaining model stability.

The adversarial transfer learning strategy shows limited effectiveness when patterns are too similar or too different. While freezing the discriminator after round 5 provides acceptable results, other rounds may yield better performance, though optimizing this parameter remains challenging due to adversarial methods' inherent limitations. Fortunately, this strategy in METR-LA still provides the most significant improvement compared to D3MGM and DPM.

\subsection{Pattern Confusion}

We present the first study on the generalizability of spatio-temporal kriging, focusing on pattern confusion through controlled tests.

\begin{figure}[t]
	\includegraphics[width=0.48\textwidth]{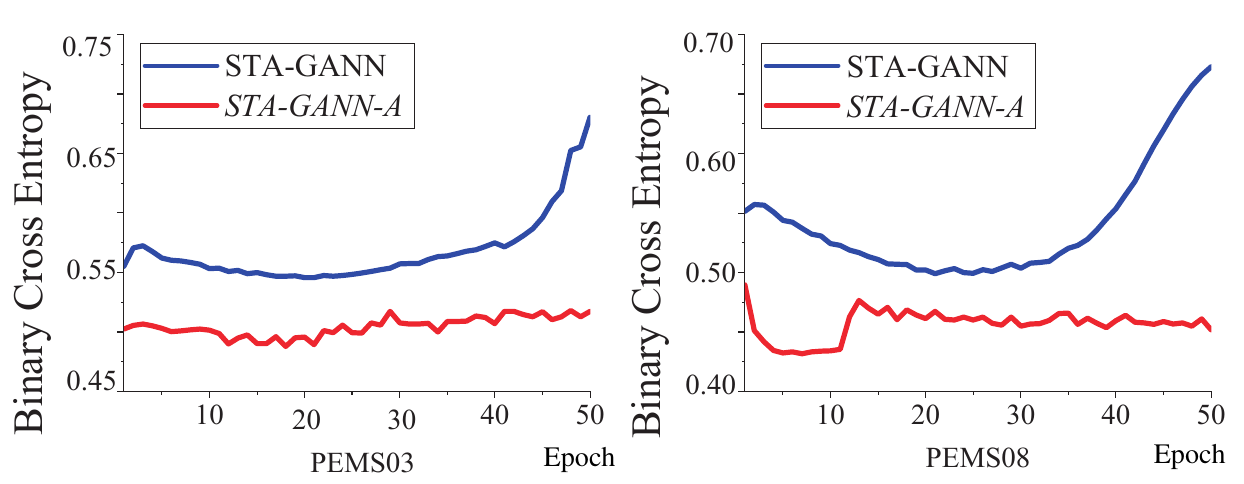}
	\vspace{-10pt}
	\caption{Confusion between known and unknown sensors on PEMS03 and PEMS08}
	\label{fusion2}
	\vspace{-10pt}
\end{figure}

To address the fundamental challenge of ensuring generalizability between sensors, we employ the adversarial transfer learning strategy, introducing a discriminator with a gradient reversal layer to enforce domain invariance. To evaluate its effect, we test STA-GANN against \textit{STA-GANN-A} to evaluate generalizability,  , which includes all components except adversarial training. We incorporate an unlearnable discriminator into \textit{STA-GANN-A} to detect the classification loss on the PEMS03 and PEMS08 test sets in Figure ~\ref{fusion2}.

The \textit{BCE} loss of \textit{STA-GANN-A} remains smooth, with a brief initial drop on PEMS08 due to overly unique patterns learned from known sensors. In contrast, STA-GANN’s loss decreases slowly at first but rises sharply after the discriminator is frozen, implying that the model maintains domain invariance between sensors.

In STA-GANN, the discriminator learns to distinguish the domains of sensors, while the gradient reversal layer reverses the update direction, preventing reliable distinction between known and unknown sensors. It forces the encoder to confuse sensor domains, leading to indistinguishable domains at later stages and the BCE loss increase—evidence of successful generalization.

In summary, an increase in BCE loss on the test set signals that the adversarial strategy has masked sensor-specific features, enabling STA-GANN to generalize from known to unknown sensors, validating both our motivations and the effectiveness of the proposed method.

\subsection{Missing Rate Experiment}

\begin{figure}[h]
	\centering
	\includegraphics[width=0.46\textwidth]{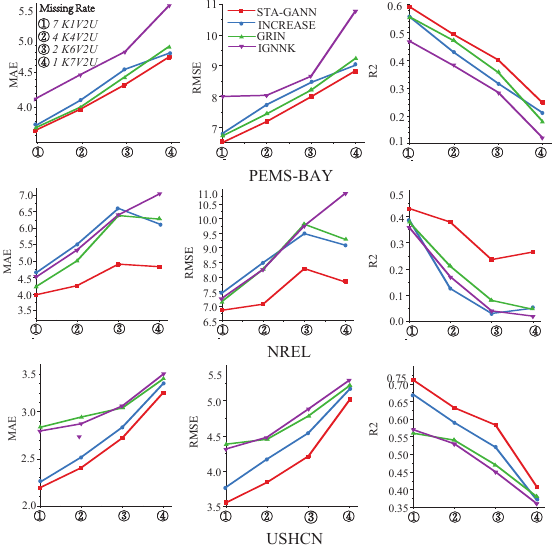}
	\vspace{-10pt}
	\caption{Experimental results for the missing rate}
	\label{MISS}
\end{figure}

Missing rate of sensors is a critical issue in spatio-temporal kriging, and we conduct experiments to simulate different missing rate scenarios in reality. In our notation, $xKyVzU$ indicates that $10\% \times x$ sensors are known, and $10\% \times z$ sensors are unknown in the test set, the  remaining sensors are unknown sensors in the validation set. Our focus lies on analyzing the spatio-temporal patterns derived from the $10\% \times x$ known sensors and evaluating the performance on the $10\% \times z$ sensors.  To ensure consistency in our evaluation, we fix $z = 2$~(representing fixed 20\% unknown sensors), and varied the number of known sensor across four configurations: $x = 7, 4, 2, 1$. For example, when $x = 1$, the known sensors must extrapolate to twice as many unknown sensors.

We conduct experiments on the PEMS-BAY, NREL, and USHCN datasets and report metrics such as MAE, RMSE, and R2 .The performance of the STA-GANN model was compared against three novel models: IGNNK, INCREASE, and GRIN. The results, presented in Figure~\ref{MISS}, reveal a consistent trend: the performance of all models deteriorates as the missingn rate increases. However, an intriguing observation emerged for the NREL dataset: the performance at $x=2$ is worse than $x=1$. This discrepancy could be attributed to the fact that additional known sensors introduced after randomization may not exhibit strong correlations with the unknown sensors, potentially hindering the propagation of temporal information.

Despite this anomaly in the NREL dataset, STA-GANN demonstrate superior robustness across all datasets. Its ability to consistently outperform other models, even as the missing rate increases, underscores its effectiveness in capturing valid spatio-temporal patterns and generalizing them to unknown sensors. This capability is particularly valuable in scenarios with sparse sensor data, where traditional methods often struggle to maintain accuracy.

\begin{figure}[h]
	\centering
	\includegraphics[width=0.47\textwidth]{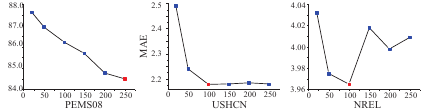}
	\vspace{-5pt}
	\caption{Experimental results for MLP embedding size}
	\label{emb}
\end{figure}
\begin{figure}[h]
	\centering
	\vspace{-15pt}
	\includegraphics[width=0.47\textwidth]{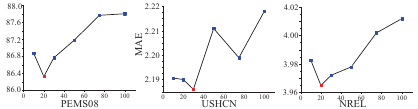}
	\vspace{-5pt}
	\caption{Experimental results for metadata embedding size}
	\label{metaemb}
\end{figure}
\begin{figure}[h]
	\centering
	\vspace{-15pt}
	\includegraphics[width=0.47\textwidth]{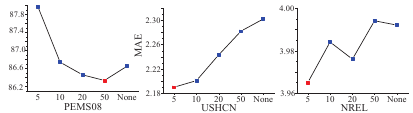}
	\vspace{-5pt}
	\caption{Experimental results for the K in adjacency matrix}
	\label{K}
\end{figure}

\begin{table*}[h]
	\caption{Runtime of STA-GANN and Baselines on nine real datasets}
	\label{resruntime}
	\renewcommand\arraystretch{0.85}
	\centering
	\vspace{-5pt}
	\resizebox{0.9\linewidth}{!}{
	\begin{tabular}{cp{45pt}|p{37pt}|p{37pt}|p{37pt}|p{30pt}|p{30pt}|p{30pt}|p{30pt}|p{30pt}}
		\toprule[1.5pt]
		Runtime(hours) & STA-GANN&GRIN&INCREASE&DualSTN&IAGCN&IGNNK&SATCN&GCN&GIN \\
		\midrule
		\multicolumn{1}{c|}{\multirow{1}{*}{METR-LA}}   &0.75&2.22&3.48&0.95&5.2&0.4&0.525&0.15&0.18\\
		
		\midrule
		\multicolumn{1}{c|}{\multirow{1}{*}{PEMS-BAY}} &1.04&3.66&4.16&1.73&10.37&0.25&0.46&0.21&0.24\\
		
		\midrule
		\multicolumn{1}{c|}{\multirow{1}{*}{PEMS03}} &0.44&1.89&3.72&1.1&8.2&0.21&0.25&0.18&0.21\\
		
		\midrule
		\multicolumn{1}{c|}{\multirow{1}{*}{PEMS04}} &0.30&1.30&2.71&0.7&3.7&0.24&0.28&0.17&0.17\\
		
		\midrule
		\multicolumn{1}{c|}{\multirow{1}{*}{PEMS07}} &1.70&7.03&12.05&3.83&20.35&0.48&0.79&0.36&0.42\\
		
		\midrule
		\multicolumn{1}{c|}{\multirow{1}{*}{PEMS08}}&0.33&0.92&1.72&0.5&2.62&0.13&0.12&0.10&0.10\\

		\midrule
		\multicolumn{1}{c|}{\multirow{1}{*}{NREL}} &1.67&4.73&8.04&2.12&12.5&0.37&0.59&0.32&0.41\\
		
		\midrule
		\multicolumn{1}{c|}{\multirow{1}{*}{USHCN}} &0.23&0.51&0.90&0.72&3.2&0.16&0.21&0.08&0.10\\
		
		\midrule
		\multicolumn{1}{c|}{\multirow{1}{*}{AQI}} &1.38&4.57&7.89&2.47&18.25&0.30&0.37&0.28&0.32\\
		
		\bottomrule[1.5pt]
	\end{tabular}}
	
\end{table*}

\subsection{Parameter study} 
For temporal embeddings, as shown in Figure~\ref{emb}, performance improves with larger dimensions but shows diminishing returns beyond 100, with potential overfitting on the NREL dataset. Metadata embedding dimensionality also requires careful tuning in Figure~\ref{metaemb}. While larger embeddings may lead to overfitting and degraded performance, a range of 20–30 typically balances model capacity and generalization effectively. Regarding the TopK parameter in the adjacency matrix, sensitivity varies across datasets as demonstrated in Figure~\ref{K}. PEMS08 and NREL show stable performance across different TopK values, while USHCN demonstrates higher sensitivity, likely due to geographic and climatic differences across regions within the United States.

\vspace{-5pt}
\subsection{Runtime experiment}\label{rt}
We record the complete runtime of the training, validation, and testing periods of all the models in the main experiment, runtime in hours is shown in Table~\ref{resruntime}.

We can find that STA-GANN is in an intermediate runtime performance. GRIN and INCREASE, which have the highest accuracy performance among the extrapolation baselines, are both at very high runtimes, while we achieve the best accuracy and keep the runtimes much lower than them. IGNNK, SATCN, and the two graph networks GCN and GIN are both shorter but have correspondingly much lower accuracies.

In short, the computational complexity of STA-GANN arises from the GIN network ($O(N H^2)$) and D3MGM ($O(N^2 E)$). GIN's complexity is slightly higher than that of GCN due to the additional MLP layers, while D3MGM's complexity is comparable to that in typical spatio-temporal forecasting tasks. In comparison, the computational cost of GRIN and INCREASE mainly comes from timestep-based GCN ($O(N^2 H L)$), while IAGCN incurs additional overhead due to graph decomposition and learning ($O(N^3 E)$).

In summary, STA-GANN not only has the best accuracy for extrapolation, the running time is much lower than the current best Baselines and is not significantly higher than the other models.

\vspace{-5pt}
\subsection{Time Shift visualization}\label{tsv}
\begin{figure}[htpb]
	\centering
	\includegraphics[width=0.46\textwidth]{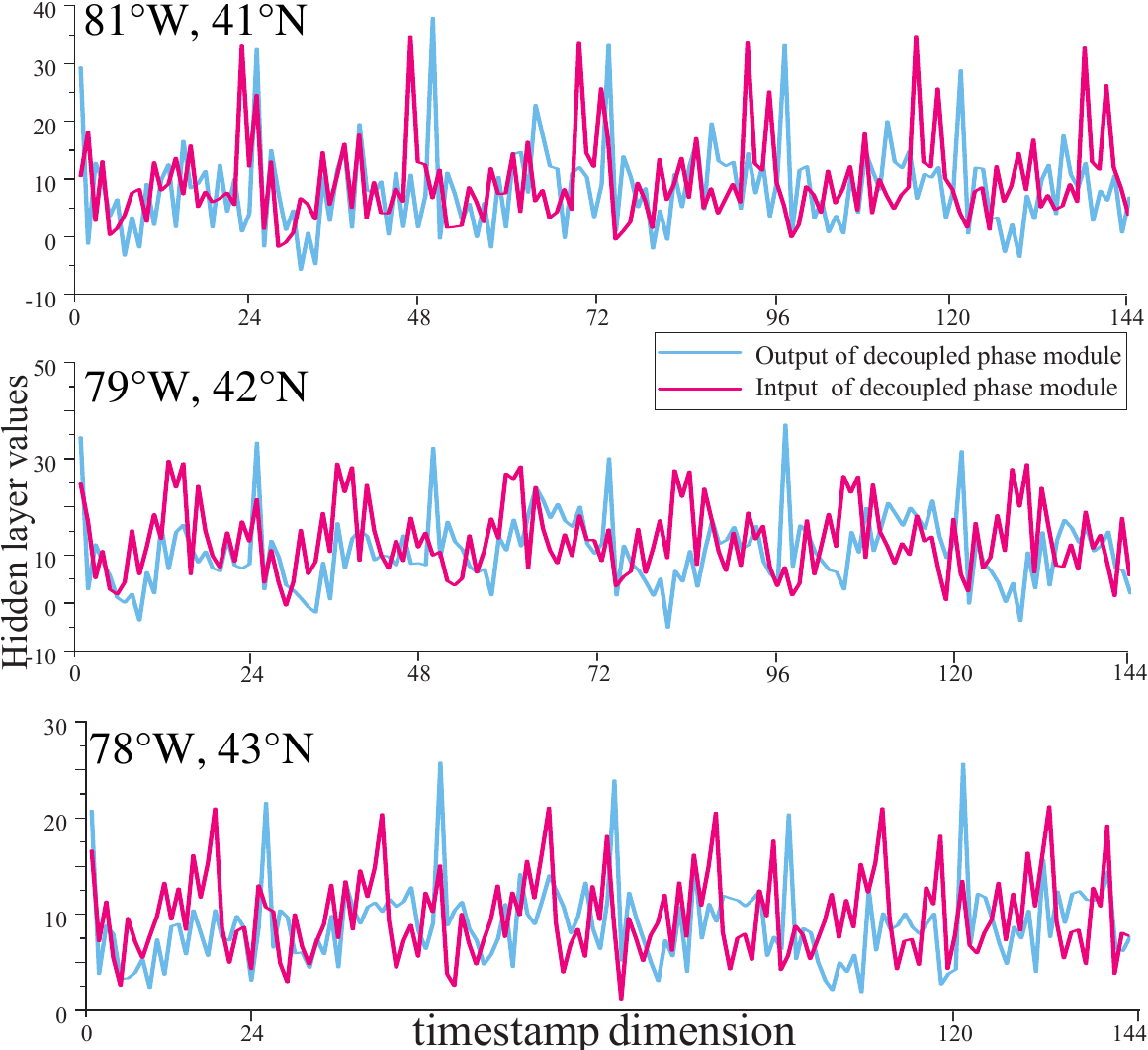}
	\caption{Outputs and inputs of Decoupled Phase Module}

	\label{dpm}
\end{figure}

We design the DPM module to learn timestamp shifts. To visualize the effect of this module, we select three sensors in USHCN and show their time series at the input and output of the DPM in Figure ~\ref{dpm}. We can observe that instead of learning the representation of time series, the DPM tries to change the phase, realizing the problem of translating the series in the timestamp dimension to accommodate the shifts, which are not exactly the same shape since we restrict only the phase of the primary frequency can be modified.

\section{Limitation}
\textit{Metadata Dependency:} Performance degrades with missing or inaccurate coordinates or timestamp shifts.

\textit{Scalability:} While node-scalable, the model falters on: (1) coordinate extrapolation in small datasets, (2) fixed-length constraints hindering real-time kriging.

\textit{Range Variation:} Failure to adapt to unknown sensors' value ranges causes inference discrepancies (Table~\ref{result}).

\section{Conclusion}
In this paper, we address the fundamental challenges of spatio-temporal kriging, particularly focusing on validity and generalization. To overcome these challenges, we present our Spatio-Temporal Aware Graph Adversarial Neural Network~(STA-GANN).
Our key contributions include threefold: First, we introduce a decoupled phase module to effectively recover timestamp shifts among sensors, ensuring the temporal validity of patterns. Second, we propose a dynamic data-driven metadata graph modeling approach to capture spatial dependencies among sensors, thereby establishing spatial pattern validity. Third, for generalizability, we employ an adversarial transfer learning strategy that enables the effective extrapolation of spatio-temporal patterns from known to unknown sensors.

Considering the wide range of real-world applications of spatio-temporal kriging, we hope that our work will inspire further investigation and advancements in this field. In the future, we will continue to focus on spatio-temporal kriging, including improving the scalability of online kriging, more sophisticated nonlinear timestamp drift and magnitude forecasting, and maintaining the Benchmark for spatio-temporal kriging.

\section{Acknowledgments}
This work is supported by NSFC No.62372430, the Youth Innovation Promotion Association CAS No.2023112, the Postdoctoral Fellowship Program of CPSF under Grant Number GZC20251078, the China Postdoctoral Science Foundation No.2025M771542 and HUA Innovation fundings. We thank all the anonymous reviewers who generously contributed their time and efforts.

\section*{GenAl Usage Disclosure}
The proposed methodology, experiments, the datasets used and the code in our paper do not use GenAI. We adopt ChatGPT to edit and improve the quality of existing text and strictly prohibit it from searching web content.

\bibliographystyle{ACM-Reference-Format}
\bibliography{reference}

\end{document}